%% file: main.tex
\pdfoutput=1

\documentclass[11pt]{article}

\usepackage[final]{acl}

\usepackage{times}
\usepackage{latexsym}

\usepackage[T1]{fontenc}

\usepackage[utf8]{inputenc}

\usepackage{microtype}

\usepackage{inconsolata}

\usepackage{graphicx}

\usepackage{enumitem}
\usepackage{geometry}

\usepackage{mathtools} 
\DeclareMathOperator*{\median}{median}
\newcommand{\Q}[2]{Q^{(#1)}_{#2}} 
\DeclarePairedDelimiter\Set\{\}{}
\DeclarePairedDelimiter\Paren()   

\usepackage{colortbl}
\usepackage{pifont}
\usepackage{booktabs}
\usepackage{amsmath}
\usepackage{cleveref}

\newcommand{\hlc}[2][yellow]{{%
    \colorlet{foo}{#1}%
    \sethlcolor{foo}\hl{#2}}%
}

\usepackage[textsize=footnotesize]{todonotes}

\usepackage[normalem]{ulem} 
\usepackage{xcolor}

\usepackage{soul}
\usepackage{booktabs}
\usepackage{tabularx}
\usepackage[table]{xcolor}

\usepackage{pifont}   
\newcommand{\checkmark}{\textcolor{green!60!black}{\ding{51}}} 
\newcommand{\crossmark}{\textcolor{red}{\ding{55}}}          
\newcommand{\unknown}{\textbf{?}}
\usepackage{fontawesome5}

\usepackage{amssymb} 
\newcommand{\official}{\(\blacktriangle\)\,}

\newcolumntype{Y}{>{\raggedleft\arraybackslash}X}
\newcommand{\auto}{\textsc{AutoRank}}

\title{\underline{\textit{Preliminary}} Ranking of WMT25 General Machine Translation Systems}

\author{
  \null \AND
  Tom Kocmi  
  \And
  Eleftherios Avramidis 
  \And
  Rachel Bawden 
  \And
  Ond\v{r}ej Bojar
  \And
  Konstantin Dranch
  \AND
  Anton Dvorkovich 
  \And
  Sergey Dukanov 
  \And
  Natalia Fedorova 
  \And
  Mark Fishel  
  \And
  Markus Freitag
  \AND
  Thamme Gowda
  \And
  Roman Grundkiewicz
  \And
  Barry Haddow  
  \And
  Marzena Karpinska 
  \AND
  Philipp Koehn 
  \And
  Howard Lakougna
  \And
  Jessica Lundin
  \And
  Kenton Murray 
  \And
  Masaaki Nagata
  \AND
  Stefano Perrella 
  \And
  Lorenzo Proietti
  \And
  Martin Popel 
  \And 
  Maja Popovi\'{c}  
  \And
  Parker Riley 
  \AND
  Mariya Shmatova 
  \And
  Stein\th\'{o}r Steingr\'{i}msson 
  \And
  Lisa Yankovskaya 
  \And 
  Vilém Zouhar
  \vspace{2cm}
}

\begin{document}
\maketitle

\section*{Introduction}

We present the \textbf{preliminary} rankings of machine translation (MT) systems submitted to the WMT25 General Machine Translation Shared Task,\footnote{\href{https://www2.statmt.org/wmt25/translation-task.html}{www2.statmt.org/wmt25/translation-task.html}} as determined by automatic evaluation metrics. Because these rankings are derived from automatic evaluation, they may exhibit a bias toward systems that employ re-ranking techniques, such as Quality Estimation or Minimum Bayes Risk decoding. The official WMT25 ranking will be based on human evaluation, which is more reliable and will supersede these results. The official WMT25 ranking will be based on human evaluation, which is more reliable and will supersede these results. The purpose of releasing these findings now is to assist task participants with their system description papers; \textit{not} to provide final findings.

\section*{Types of Systems}

We distinguish two types of MT systems participating in the shared task:
\begin{itemize}
\item \textbf{Constrained systems:} must use only publicly available training data and models, be limited to a maximum of 20B parameters, and have their model weights released under an open license.

\item \textbf{Unconstrained systems:} (marked with \hlc[gray!30]{gray}) are all other systems. They have no restrictions on training data or model size, and there is no requirement to publish the model weights. This category also includes systems for which training information is not public.

\end{itemize}

\begin{table}[t]
\centering
\small
\setlength{\tabcolsep}{5pt}
\begin{tabular}{l r c}
\toprule
\textbf{Model} & \textbf{\# Params} & \textbf{Open?} \\
\midrule
\multicolumn{3}{l}{\textsc{\textbf{\color{purple} Constrained systems}}}\\
AyaExpanse-8B        & 8B                 & \faCheck \\
CommandR7B           & 7B                 & \faCheck \\
EuroLLM-9B           & 9B                 & \faCheck \\
Gemma-3-12B          & 12B                & \faCheck \\
Llama-3.1-8B         & 8B                 & \faCheck \\
Mistral-7B           & 7.3B               & \faCheck \\
NLLB (NLLB-200)      & 3.3B               & \faCheck \\
Qwen2.5-7B           & 7.6B               & \faCheck \\
TowerPlus-9B         & 9B                 & \faCheck \\
\midrule
\multicolumn{3}{l}{\textsc{\textbf{\color{teal} Unconstrained systems}}}\\
AyaExpanse-32B       & 32B                & \faCheck \\
Claude-4             & ---                & \faTimes \\
CommandA             & 111B               & \faCheck \\
DeepSeek-V3          & 671B (37B act.)    & \faCheck \\
EuroLLM-22B          & 22B (preview)      & \faCheck \\
Gemma-3-27B          & 27B                & \faCheck \\
Gemini-2.5-Pro       & ---                & \faTimes \\
GPT-4.1              & ---                & \faTimes \\
Llama-4-Maverick     & ---                & \faCheck \\
Mistral-Medium       & ---                & \faTimes \\
ONLINE-B             & ---                & \faTimes \\
ONLINE-G             & ---                & \faTimes \\
ONLINE-W             & ---                & \faTimes \\
Qwen3-235B           & 235B (22B act.)    & \faCheck \\
TowerPlus-72B        & 72B                & \faCheck \\
\midrule
\end{tabular}
\caption{List of constrained and unconstrained systems with parameter count. Open-weight models were marked with a checkmark (\faCheck).}
\label{tab:models}
\end{table}

\section*{Evaluated Systems}

\paragraph{Models.} Our evaluation includes systems submitted by participants, as well as open-weight and proprietary models. We selected the largest or best-performing version of each model where applicable. All constrained and unconstrained systems are listed in \autoref{tab:models}. Full details for all systems will be available in the upcoming WMT25 finding paper.



\paragraph{Prompts.} We prompt all language models using a zero-shot, instruction-following approach, with the specific instructions provided as part of the blind test set. Each model was first tasked with translating the entire document. If this initial attempt failed (e.g., due to producing an incorrect paragraph count or exceeding the token limit), we implemented a fallback strategy of translating the document paragraph by paragraph. This generic setup may disadvantage systems tuned for specific MT instructions, such as TowerLLM or EuroLLM; these are marked with~[M]. 


Additionally, we made two model-specific adjustments: (1) for \textbf{Qwen3-235B} reasoning capabilities were disabled, and (2) for \textbf{Gemini-2.5-Pro}, no reasoning budget was set, which resulted in a 6.6$\times$ increase in output tokens, making it the most expensive model to evaluate.


The code for collecting translations is publicly available at \faGithub\ \href{https://github.com/wmt-conference/wmt-collect-translations}{github.com/wmt-conference/wmt-collect-translations} and we marked all systems collected by us with \official{}.

\section*{Evaluation Data}

\paragraph{Languages.} The evaluation covers 32 language pairs, with each test set containing approximately 37K words\footnote{As per whitespace split.} organized into documents. Half of these language pairs are designated for human evaluation, while the other half belong to a multilingual subtrack evaluated solely with automatic metrics. Most pairs are in the English-to-X direction with the exception of Czech-Ukrainian, Czech-German, and Japanese-Chinese.

\paragraph{Domains.} The test sets combine material from four distinct domains, though language pairs with a non-English source may differ slightly in their domain distribution and dataset size. The domains include:
\begin{itemize}
    \item \textbf{News commentary}
    \item \textbf{Social} (texts from social networks, collected with screenshots)
    \item \textbf{Speech} (transcripts of speeches obtained automatically)
    \item \textbf{Literary} (two documents of roughly 5{,}000 words each)
\end{itemize}

\paragraph{Data preprocessing.} Each document is divided into segments of approximately 100 words. These segments typically correspond to natural paragraphs; however, some paragraphs were split (or merged) to adhere to this length constraint. Sentence splitting was not performed on the source texts, so segments often contain multiple sentences. For the multimodal context, participants were allowed to (optionally) use accompanying image and video modalities when available, and human annotators who provided reference translations also had access to this context.

All data for this task, including references, system outputs, automatic scores, and the LaTeX source of this document, are publicly available at our repository: \href{https://github.com/wmt-conference/wmt25-general-mt}{github.com/wmt-conference/wmt25-general-mt}.




\section*{Automatic Ranking}
\label{sec:automatic-ranking}

This section details our automatic ranking method, which we refer to as \auto. 
Both the set of automatic metrics and the aggregation procedure have been slightly updated since last year's shared task.

\paragraph{Metrics.}
For most language pairs,\footnote{See ``Low-resource exception'' below.} the \auto\ is a combination of three distinct families of evaluation methods:
\begin{itemize}
    \item \textbf{LLM-as-a-Judge (reference-less).} We use GEMBA-ESA \citep{kocmi-federmann-2023-large} with two independent judges: GPT-4.1 \citep{openai_gpt41_2025} and Command A \citep{cohere2025commandaenterprisereadylarge}, both in a reference-less setting.
    \item \textbf{Trained reference-based metrics.} Two supervised metrics trained to approximate human quality judgments with references: MetricX-24-Hybrid-XL \citep{juraska-etal-2024-metricx} and XCOMET-XL \citep{guerreiro-etal-2024-xcomet}.
    \item \textbf{Trained Quality Estimation (QE).} The reference-less QE metric CometKiwi-XL \citep{rei-etal-2023-scaling}, which is also trained to mimic human judgments.
\end{itemize}



This combination of reference-based and reference-less (or QE) methods is designed to balance their complementary failure modes. Reference-based metrics typically achieve a higher correlation with human judgments when high-quality references are available, while reference-less methods reduce susceptibility to reference bias when references are suboptimal \citep{freitag-etal-2023-results}. We also account for known issues with specific metrics. To mitigate a common QE pitfall, i.e., being fooled by fluent output in the wrong language, the GEMBA-ESA prompt explicitly specifies the target language. However, while GEMBA-ESA is intended to reduce bias toward systems that use re-ranking, we note that some participants incorporated it directly as a reward model.

\paragraph{System-level scores.}
The system-level score for each language pair is the average of its paragraph-level (segment-level) scores from each metric across the testset. We make one exception for language pairs without human references by excluding CometKiwi-XL from the \auto\ computation. This avoids redundancy, as the other hybrid metrics (MetricX-24-Hybrid-XL and XCOMET-XL) can also run in a reference-less (QE) mode to provide the necessary QE signal.

\paragraph{Low-resource exception.}
For the two lowest-resource languages in the testset, i.e., \textbf{Bhojpuri} and \textbf{Maasai}, we rely solely on \texttt{chrF++} \citep{popovic-2017-chrf}, computed with \texttt{sacrebleu} \citep{post-2018-call}. This approach was chosen because the reliability of our main metrics is unestablished for these languages \citep{falcao-etal-2024-comet,singh-etal-2024-good,wang-etal-2024-evaluating,sindhujan-etal-2025-llms}, whereas high-quality human references required for \texttt{chrF++} were available.

\paragraph{From system-level scores to \auto.}
To combine the metrics into a single score, we first normalize them using median-interpercentile scaling to address differences in scale and reduce the influence of low-performing outliers.
We then compute the average using equal weights. Finally, we linearly rescale the results to the range from 1 to $N$ systems. A detailed description is provided below:

Let $S$ be the set of submitted systems for a given language pair, $|S|=N$, and let $M$ be the set of automatic metrics used for that language pair (for Bhojpuri and Maasai, $|M|=1$). For each metric $m\in M$ and system $s\in S$, we compute a system-level score $x^{(m)}_s$ as the average of that metric over all available test segments. To combine scores across metrics, we first map them to a common scale; however, classical min–max normalization is highly sensitive to outliers. To downweight extremes without discarding any system, we apply a \emph{median--inter\-percentile} scaling to each metric $m$:
\begin{subequations}\label{eq:robust}
\begin{align}
\tilde{x}^{(m)} &= \median\,\Set*{x^{(m)}_s \mid s \in S},\\[2pt]
D^{(m)} &= \max\,\Paren*{\varepsilon,\, \Q{m}{100} - \Q{m}{25}},\\[2pt]
z^{(m)}_s &= \frac{x^{(m)}_s - \tilde{x}^{(m)}}{D^{(m)}}.
\end{align}
\end{subequations}
Where $\varepsilon>0$ and $\Q{m}{p}$ denotes the $p$-th percentile of $\{x^{(m)}_s : s\in S\}$. Importantly, Eq.~\eqref{eq:robust} is continuous and monotonic: it keeps all systems and preserves their order within each metric. Then, for each system, we average the robust-scaled values across metrics:
\begin{equation}
    \bar{z}_s \;=\; \frac{1}{|M|}\sum_{m\in M} z^{(m)}_s .
    \label{eq:avgz}
\end{equation}
Averaging after robust scaling yields a single comparable score that preserves the magnitude of performance differences between systems (in standardized units) while preventing any single metric’s outliers from dominating. Finally, for readability and to follow the WMT convention from last year (lower is better in \auto, i.e., $1$ is best and $N$ worst), we apply a final linear mapping to the set $\{\bar{z}_s\}_{s\in S}$. Specifically, within $\{\bar{z}_s\}_{s\in S}$ the system with the highest average score is assigned $1$, the system with the lowest average score is assigned $N$, and all remaining systems are placed linearly between these two endpoints. This remapping is applied only once—after the cross-metric aggregation—so it preserves the ordering and relative spacing between systems while retaining the outlier mitigation provided by the robust scaling. We refer to the resulting value as \auto\ in the various tables.

\section*{Human Evaluation}

This year's shared task saw a record number of participants, with submissions from 36 unique teams.\footnote{A total of 43 teams initially registered, but 7 later withdrew or were disqualified.} Due to budget constraints, we could not include all submissions in the human evaluation. Therefore, we selected a subset of systems for manual evaluation using the Error Span Annotation (ESA) protocol \citep{kocmi2024errorspanannotationbalanced}. This subset typically consists of 18 systems per language pair, although this number is higher for some. For any system not selected for human evaluation, the automatic metric ranking (\auto) serves as the official final result.

The selection process for human evaluation prioritized \textbf{constrained} systems over \textbf{unconstrained} ones, following a two-step procedure:
\begin{enumerate}
    \item First, the top-8 performing \textbf{constrained} systems were selected, based on their automatic scores.
    \item Second, the overall top-performing systems from the remaining pool (both constrained and unconstrained) were added until a total of 18 systems was reached.
\end{enumerate}




\section*{Limitations}

A key limitation of our evaluation is that some models have been optimized for the very metrics we employ in \auto, either during training or at inference time \cite{freitag-etal-2022-high, finkelstein2024mbr}. This can lead to artificially inflated scores that do not accurately reflect a model's true capabilities \cite{kovacs-etal-2024-mitigating}. To mitigate this issue, we aggregate the assessments from multiple learned metrics and LLM-as-a-judge approaches. However, even this strategy has shortcomings. First, scores from different learned metrics often exhibit high correlation among themselves. Second, LLM-as-a-judge approaches, including Gemba-ESA, may themselves have been leveraged to optimize machine translation models.

Another limitation is that we use automatic metrics to evaluate entire paragraphs, whereas their reliability is typically established at the sentence level. Additionally, learned metrics struggle with low-resource languages, such as English-to-Bhojpuri and English-to-Maasai. For these cases, we rely on \texttt{chrF++} instead. However, \texttt{chrF++} is a surface-level metric that, like BLEU, has repeatedly been shown to correlate poorly with human judgments \citep{kocmi-etal-2021-ship, freitag-etal-2022-results, freitag-etal-2023-results}.

Furthermore, our automatic evaluation is conducted at the paragraph level, without incorporating document-level context. This may lead to inflated scores for systems that translate the dataset paragraph by paragraph, disregarding dependencies and coherence across paragraphs.

The LLM-as-a-judge approach also depends on the language performance of the underlying LLMs. For our evaluation, we selected two top-performing multilingual systems: GPT-4.1 and Command A. Command A officially supports only 23 languages \citep{cohere2025commandaenterprisereadylarge}, while the set of languages supported by GPT-4.1 is not publicly documented. Nevertheless, as both metrics correlate well across all languages and show strong agreement with other evaluation metrics, we retained them as judges for all 30 language pairs.

Finally, using automatically generated speech recognition transcripts as source text in the speech domain introduces additional noise, as the evaluation metrics are unlikely to be robust to ASR errors. Consequently, systems that handle the speech domain well may receive lower scores if their outputs diverge from the ASR transcript, even when their translations are correct.

Given these challenges, along with the well-documented biases and limitations of automatic metrics \citep{karpinska-etal-2022-demetr, moghe2024machine}, human evaluation remains indispensable. Accordingly, results from human assessments will supersede the automatic rankings presented here.

\section*{Acknowledgement}
This report would not have been possible without the partnership with Árni Magnússon Institute for Icelandic Studies, Charles University, Cohere, Custom.MT, Dubformer, Gates Foundation, Google, Institute of the Estonian Language, Microsoft, NTT, Toloka, University of Tartu, University of Tokyo.
Furthermore, we are grateful to Toshiaki Nakazawa.

\newcolumntype{L}[1]{>{\raggedright\let\newline\\\arraybackslash\hspace{0pt}}m{#1}}
\newcolumntype{C}[1]{>{\centering\let\newline\\\arraybackslash\hspace{0pt}}m{#1}}
\newcolumntype{R}[1]{>{\raggedleft\let\newline\\\arraybackslash\hspace{0pt}}m{#1}}

\input{generated_report.tex}

\clearpage
\clearpage

\bibliography{anthology.min.bib,custom}

\appendix
\onecolumn

\section{Metrics correlations}

To examine how the metrics used for \auto correlate with each other, we calculated the Pearson correlation between paragraph-level scores for all systems, resulting in a sample size of around 14k scores per each language pair.

The results show that GEMBA-ESA on CmdA and GPT-4.1 exhibit the highest correlations for almost all languages. In contrast, the weakest correlations are generally observed between xComet and both GEMBA-ESA variants.

When examining results by language pair, Bhojpuri, Maasai, and Marathi show the lowest correlations. This is why we use \texttt{chrF++} for the first two language pairs. Unfortunately, no reference translations are available for Marathi, so we must rely on QE metrics for its evaluation.

\begin{table*}[h]
\small
\centering
\setlength{\tabcolsep}{4.9pt}
\input{metrics_correlations}
\end{table*}

\end{document}

%% file: metrics_correlations.tex
\begin{tabular}{lllllllllll}
\toprule
 & \shortstack{Kiwi\\G-CmdA} & \shortstack{Kiwi\\G-GPT} & \shortstack{Kiwi\\MetX} & \shortstack{Kiwi\\xComet} & \shortstack{G-CmdA\\G-GPT} & \shortstack{G-CmdA\\MetX} & \shortstack{G-CmdA\\xComet} & \shortstack{G-GPT\\MetX} & \shortstack{G-GPT\\xComet} & \shortstack{MetX\\xComet} \\
\midrule
cs-de\_DE & \cellcolor[HTML]{E9F6A1}\textcolor[HTML]{000000}{0.441} & \cellcolor[HTML]{D9EF8B}\textcolor[HTML]{000000}{0.484} & \cellcolor[HTML]{BBE278}\textcolor[HTML]{000000}{0.541} & \cellcolor[HTML]{48AE5C}\textcolor[HTML]{000000}{0.709} & \cellcolor[HTML]{36A657}\textcolor[HTML]{FFFFFF}{0.732} & \cellcolor[HTML]{A2D76A}\textcolor[HTML]{000000}{0.583} & \cellcolor[HTML]{FAFDB8}\textcolor[HTML]{000000}{0.403} & \cellcolor[HTML]{7FC866}\textcolor[HTML]{000000}{0.636} & \cellcolor[HTML]{ECF7A6}\textcolor[HTML]{000000}{0.436} & \cellcolor[HTML]{AFDD70}\textcolor[HTML]{000000}{0.560} \\
cs-uk\_UA & \cellcolor[HTML]{BFE47A}\textcolor[HTML]{000000}{0.531} & \cellcolor[HTML]{98D368}\textcolor[HTML]{000000}{0.600} & \cellcolor[HTML]{54B45F}\textcolor[HTML]{000000}{0.696} & \cellcolor[HTML]{128A49}\textcolor[HTML]{FFFFFF}{0.794} & \cellcolor[HTML]{48AE5C}\textcolor[HTML]{000000}{0.708} & \cellcolor[HTML]{A9DA6C}\textcolor[HTML]{000000}{0.571} & \cellcolor[HTML]{C7E77F}\textcolor[HTML]{000000}{0.517} & \cellcolor[HTML]{73C264}\textcolor[HTML]{000000}{0.654} & \cellcolor[HTML]{A9DA6C}\textcolor[HTML]{000000}{0.573} & \cellcolor[HTML]{48AE5C}\textcolor[HTML]{000000}{0.710} \\
en-ar\_EG & \cellcolor[HTML]{91D068}\textcolor[HTML]{000000}{0.610} & \cellcolor[HTML]{A9DA6C}\textcolor[HTML]{000000}{0.573} & \cellcolor[HTML]{279F53}\textcolor[HTML]{FFFFFF}{0.750} & \cellcolor[HTML]{D3EC87}\textcolor[HTML]{000000}{0.494} & \cellcolor[HTML]{30A356}\textcolor[HTML]{FFFFFF}{0.740} & \cellcolor[HTML]{87CB67}\textcolor[HTML]{000000}{0.624} & \cellcolor[HTML]{FFF2AA}\textcolor[HTML]{000000}{0.350} & \cellcolor[HTML]{93D168}\textcolor[HTML]{000000}{0.605} & \cellcolor[HTML]{FED27F}\textcolor[HTML]{000000}{0.268} & \cellcolor[HTML]{ADDC6F}\textcolor[HTML]{000000}{0.565} \\
en-bho\_IN & \cellcolor[HTML]{E0F295}\textcolor[HTML]{000000}{0.465} & \cellcolor[HTML]{F16640}\textcolor[HTML]{000000}{0.093} & \cellcolor[HTML]{C7E77F}\textcolor[HTML]{000000}{0.517} & \cellcolor[HTML]{DD3D2D}\textcolor[HTML]{FFFFFF}{0.030} & \cellcolor[HTML]{CFEB85}\textcolor[HTML]{000000}{0.503} & \cellcolor[HTML]{89CC67}\textcolor[HTML]{000000}{0.621} & \cellcolor[HTML]{E34933}\textcolor[HTML]{FFFFFF}{0.051} & \cellcolor[HTML]{EFF8AA}\textcolor[HTML]{000000}{0.428} & \cellcolor[HTML]{CC2627}\textcolor[HTML]{FFFFFF}{-0.008} & \cellcolor[HTML]{FCAA5F}\textcolor[HTML]{000000}{0.194} \\
en-bn\_BD & \cellcolor[HTML]{2DA155}\textcolor[HTML]{FFFFFF}{0.742} & \cellcolor[HTML]{279F53}\textcolor[HTML]{FFFFFF}{0.752} & \cellcolor[HTML]{0B7D42}\textcolor[HTML]{FFFFFF}{0.822} & \cellcolor[HTML]{D1EC86}\textcolor[HTML]{000000}{0.498} & \cellcolor[HTML]{108647}\textcolor[HTML]{FFFFFF}{0.802} & \cellcolor[HTML]{33A456}\textcolor[HTML]{FFFFFF}{0.735} & \cellcolor[HTML]{ECF7A6}\textcolor[HTML]{000000}{0.435} & \cellcolor[HTML]{36A657}\textcolor[HTML]{FFFFFF}{0.730} & \cellcolor[HTML]{E8F59F}\textcolor[HTML]{000000}{0.448} & \cellcolor[HTML]{A2D76A}\textcolor[HTML]{000000}{0.584} \\
en-cs\_CZ & \cellcolor[HTML]{8CCD67}\textcolor[HTML]{000000}{0.617} & \cellcolor[HTML]{54B45F}\textcolor[HTML]{000000}{0.696} & \cellcolor[HTML]{39A758}\textcolor[HTML]{FFFFFF}{0.728} & \cellcolor[HTML]{2AA054}\textcolor[HTML]{FFFFFF}{0.747} & \cellcolor[HTML]{219C52}\textcolor[HTML]{FFFFFF}{0.757} & \cellcolor[HTML]{7AC665}\textcolor[HTML]{000000}{0.642} & \cellcolor[HTML]{BFE47A}\textcolor[HTML]{000000}{0.533} & \cellcolor[HTML]{5DB961}\textcolor[HTML]{000000}{0.682} & \cellcolor[HTML]{BDE379}\textcolor[HTML]{000000}{0.535} & \cellcolor[HTML]{45AD5B}\textcolor[HTML]{000000}{0.712} \\
en-de\_DE & \cellcolor[HTML]{DAF08D}\textcolor[HTML]{000000}{0.481} & \cellcolor[HTML]{B7E075}\textcolor[HTML]{000000}{0.546} & \cellcolor[HTML]{8ECF67}\textcolor[HTML]{000000}{0.612} & \cellcolor[HTML]{138C4A}\textcolor[HTML]{FFFFFF}{0.789} & \cellcolor[HTML]{2DA155}\textcolor[HTML]{FFFFFF}{0.742} & \cellcolor[HTML]{A5D86A}\textcolor[HTML]{000000}{0.578} & \cellcolor[HTML]{FFF2AA}\textcolor[HTML]{000000}{0.350} & \cellcolor[HTML]{9BD469}\textcolor[HTML]{000000}{0.593} & \cellcolor[HTML]{FFF5AE}\textcolor[HTML]{000000}{0.358} & \cellcolor[HTML]{B1DE71}\textcolor[HTML]{000000}{0.559} \\
en-el\_GR & \cellcolor[HTML]{33A456}\textcolor[HTML]{FFFFFF}{0.736} & \cellcolor[HTML]{17934E}\textcolor[HTML]{FFFFFF}{0.777} & \cellcolor[HTML]{148E4B}\textcolor[HTML]{FFFFFF}{0.787} & \cellcolor[HTML]{57B65F}\textcolor[HTML]{000000}{0.691} & \cellcolor[HTML]{006837}\textcolor[HTML]{FFFFFF}{0.863} & \cellcolor[HTML]{42AC5A}\textcolor[HTML]{000000}{0.716} & \cellcolor[HTML]{B9E176}\textcolor[HTML]{000000}{0.542} & \cellcolor[HTML]{2DA155}\textcolor[HTML]{FFFFFF}{0.743} & \cellcolor[HTML]{B9E176}\textcolor[HTML]{000000}{0.544} & \cellcolor[HTML]{30A356}\textcolor[HTML]{FFFFFF}{0.741} \\
en-et\_EE & \cellcolor[HTML]{15904C}\textcolor[HTML]{FFFFFF}{0.783} & \cellcolor[HTML]{07753E}\textcolor[HTML]{FFFFFF}{0.837} & \cellcolor[HTML]{0A7B41}\textcolor[HTML]{FFFFFF}{0.825} & \cellcolor[HTML]{3FAA59}\textcolor[HTML]{000000}{0.720} & \cellcolor[HTML]{148E4B}\textcolor[HTML]{FFFFFF}{0.787} & \cellcolor[HTML]{33A456}\textcolor[HTML]{FFFFFF}{0.736} & \cellcolor[HTML]{A2D76A}\textcolor[HTML]{000000}{0.583} & \cellcolor[HTML]{128A49}\textcolor[HTML]{FFFFFF}{0.795} & \cellcolor[HTML]{73C264}\textcolor[HTML]{000000}{0.655} & \cellcolor[HTML]{108647}\textcolor[HTML]{FFFFFF}{0.802} \\
en-fa\_IR & \cellcolor[HTML]{0D8044}\textcolor[HTML]{FFFFFF}{0.814} & \cellcolor[HTML]{07753E}\textcolor[HTML]{FFFFFF}{0.834} & \cellcolor[HTML]{006837}\textcolor[HTML]{FFFFFF}{0.862} & \cellcolor[HTML]{4EB15D}\textcolor[HTML]{000000}{0.703} & \cellcolor[HTML]{036E3A}\textcolor[HTML]{FFFFFF}{0.852} & \cellcolor[HTML]{15904C}\textcolor[HTML]{FFFFFF}{0.785} & \cellcolor[HTML]{9BD469}\textcolor[HTML]{000000}{0.596} & \cellcolor[HTML]{128A49}\textcolor[HTML]{FFFFFF}{0.793} & \cellcolor[HTML]{9DD569}\textcolor[HTML]{000000}{0.589} & \cellcolor[HTML]{57B65F}\textcolor[HTML]{000000}{0.689} \\
en-hi\_IN & \cellcolor[HTML]{75C465}\textcolor[HTML]{000000}{0.651} & \cellcolor[HTML]{6EC064}\textcolor[HTML]{000000}{0.663} & \cellcolor[HTML]{73C264}\textcolor[HTML]{000000}{0.654} & \cellcolor[HTML]{E9F6A1}\textcolor[HTML]{000000}{0.443} & \cellcolor[HTML]{249D53}\textcolor[HTML]{FFFFFF}{0.754} & \cellcolor[HTML]{70C164}\textcolor[HTML]{000000}{0.658} & \cellcolor[HTML]{EEF8A8}\textcolor[HTML]{000000}{0.432} & \cellcolor[HTML]{60BA62}\textcolor[HTML]{000000}{0.681} & \cellcolor[HTML]{E3F399}\textcolor[HTML]{000000}{0.459} & \cellcolor[HTML]{7FC866}\textcolor[HTML]{000000}{0.634} \\
en-id\_ID & \cellcolor[HTML]{54B45F}\textcolor[HTML]{000000}{0.696} & \cellcolor[HTML]{17934E}\textcolor[HTML]{FFFFFF}{0.777} & \cellcolor[HTML]{4BB05C}\textcolor[HTML]{000000}{0.705} & \cellcolor[HTML]{60BA62}\textcolor[HTML]{000000}{0.680} & \cellcolor[HTML]{17934E}\textcolor[HTML]{FFFFFF}{0.775} & \cellcolor[HTML]{82C966}\textcolor[HTML]{000000}{0.633} & \cellcolor[HTML]{B9E176}\textcolor[HTML]{000000}{0.542} & \cellcolor[HTML]{73C264}\textcolor[HTML]{000000}{0.653} & \cellcolor[HTML]{B3DF72}\textcolor[HTML]{000000}{0.552} & \cellcolor[HTML]{17934E}\textcolor[HTML]{FFFFFF}{0.775} \\
en-is\_IS & \cellcolor[HTML]{148E4B}\textcolor[HTML]{FFFFFF}{0.787} & \cellcolor[HTML]{0E8245}\textcolor[HTML]{FFFFFF}{0.811} & \cellcolor[HTML]{06733D}\textcolor[HTML]{FFFFFF}{0.839} & \cellcolor[HTML]{70C164}\textcolor[HTML]{000000}{0.659} & \cellcolor[HTML]{249D53}\textcolor[HTML]{FFFFFF}{0.756} & \cellcolor[HTML]{45AD5B}\textcolor[HTML]{000000}{0.713} & \cellcolor[HTML]{D3EC87}\textcolor[HTML]{000000}{0.495} & \cellcolor[HTML]{148E4B}\textcolor[HTML]{FFFFFF}{0.787} & \cellcolor[HTML]{89CC67}\textcolor[HTML]{000000}{0.620} & \cellcolor[HTML]{30A356}\textcolor[HTML]{FFFFFF}{0.741} \\
en-it\_IT & \cellcolor[HTML]{B5DF74}\textcolor[HTML]{000000}{0.549} & \cellcolor[HTML]{9BD469}\textcolor[HTML]{000000}{0.596} & \cellcolor[HTML]{57B65F}\textcolor[HTML]{000000}{0.691} & \cellcolor[HTML]{16914D}\textcolor[HTML]{FFFFFF}{0.780} & \cellcolor[HTML]{33A456}\textcolor[HTML]{FFFFFF}{0.735} & \cellcolor[HTML]{ADDC6F}\textcolor[HTML]{000000}{0.566} & \cellcolor[HTML]{DFF293}\textcolor[HTML]{000000}{0.470} & \cellcolor[HTML]{A2D76A}\textcolor[HTML]{000000}{0.583} & \cellcolor[HTML]{E3F399}\textcolor[HTML]{000000}{0.456} & \cellcolor[HTML]{42AC5A}\textcolor[HTML]{000000}{0.716} \\
en-ja\_JP & \cellcolor[HTML]{7AC665}\textcolor[HTML]{000000}{0.644} & \cellcolor[HTML]{69BE63}\textcolor[HTML]{000000}{0.668} & \cellcolor[HTML]{42AC5A}\textcolor[HTML]{000000}{0.717} & \cellcolor[HTML]{57B65F}\textcolor[HTML]{000000}{0.691} & \cellcolor[HTML]{279F53}\textcolor[HTML]{FFFFFF}{0.752} & \cellcolor[HTML]{87CB67}\textcolor[HTML]{000000}{0.626} & \cellcolor[HTML]{B9E176}\textcolor[HTML]{000000}{0.543} & \cellcolor[HTML]{7FC866}\textcolor[HTML]{000000}{0.637} & \cellcolor[HTML]{D3EC87}\textcolor[HTML]{000000}{0.496} & \cellcolor[HTML]{42AC5A}\textcolor[HTML]{000000}{0.715} \\
en-kn\_IN & \cellcolor[HTML]{128A49}\textcolor[HTML]{FFFFFF}{0.796} & \cellcolor[HTML]{17934E}\textcolor[HTML]{FFFFFF}{0.778} & \cellcolor[HTML]{0A7B41}\textcolor[HTML]{FFFFFF}{0.826} & \cellcolor[HTML]{FFFCBA}\textcolor[HTML]{000000}{0.379} & \cellcolor[HTML]{138C4A}\textcolor[HTML]{FFFFFF}{0.790} & \cellcolor[HTML]{45AD5B}\textcolor[HTML]{000000}{0.714} & \cellcolor[HTML]{FEEA9B}\textcolor[HTML]{000000}{0.324} & \cellcolor[HTML]{4EB15D}\textcolor[HTML]{000000}{0.703} & \cellcolor[HTML]{FFFBB8}\textcolor[HTML]{000000}{0.375} & \cellcolor[HTML]{ADDC6F}\textcolor[HTML]{000000}{0.563} \\
en-ko\_KR & \cellcolor[HTML]{78C565}\textcolor[HTML]{000000}{0.645} & \cellcolor[HTML]{6BBF64}\textcolor[HTML]{000000}{0.667} & \cellcolor[HTML]{51B35E}\textcolor[HTML]{000000}{0.699} & \cellcolor[HTML]{60BA62}\textcolor[HTML]{000000}{0.680} & \cellcolor[HTML]{18954F}\textcolor[HTML]{FFFFFF}{0.774} & \cellcolor[HTML]{7AC665}\textcolor[HTML]{000000}{0.643} & \cellcolor[HTML]{A5D86A}\textcolor[HTML]{000000}{0.580} & \cellcolor[HTML]{78C565}\textcolor[HTML]{000000}{0.648} & \cellcolor[HTML]{B7E075}\textcolor[HTML]{000000}{0.547} & \cellcolor[HTML]{30A356}\textcolor[HTML]{FFFFFF}{0.738} \\
en-lt\_LT & \cellcolor[HTML]{118848}\textcolor[HTML]{FFFFFF}{0.798} & \cellcolor[HTML]{07753E}\textcolor[HTML]{FFFFFF}{0.837} & \cellcolor[HTML]{016A38}\textcolor[HTML]{FFFFFF}{0.858} & \cellcolor[HTML]{3CA959}\textcolor[HTML]{FFFFFF}{0.726} & \cellcolor[HTML]{097940}\textcolor[HTML]{FFFFFF}{0.828} & \cellcolor[HTML]{249D53}\textcolor[HTML]{FFFFFF}{0.755} & \cellcolor[HTML]{B1DE71}\textcolor[HTML]{000000}{0.556} & \cellcolor[HTML]{15904C}\textcolor[HTML]{FFFFFF}{0.783} & \cellcolor[HTML]{96D268}\textcolor[HTML]{000000}{0.601} & \cellcolor[HTML]{1E9A51}\textcolor[HTML]{FFFFFF}{0.762} \\
en-mas\_KE & \cellcolor[HTML]{54B45F}\textcolor[HTML]{000000}{0.694} & \cellcolor[HTML]{FEEA9B}\textcolor[HTML]{000000}{0.325} & \cellcolor[HTML]{FAFDB8}\textcolor[HTML]{000000}{0.403} & \cellcolor[HTML]{F67A49}\textcolor[HTML]{000000}{0.124} & \cellcolor[HTML]{E2F397}\textcolor[HTML]{000000}{0.460} & \cellcolor[HTML]{F8FCB6}\textcolor[HTML]{000000}{0.406} & \cellcolor[HTML]{FDBB6C}\textcolor[HTML]{000000}{0.223} & \cellcolor[HTML]{F16640}\textcolor[HTML]{000000}{0.096} & \cellcolor[HTML]{A50026}\textcolor[HTML]{FFFFFF}{-0.085} & \cellcolor[HTML]{BFE47A}\textcolor[HTML]{000000}{0.533} \\
en-mr\_IN & \cellcolor[HTML]{30A356}\textcolor[HTML]{FFFFFF}{0.738} & \cellcolor[HTML]{89CC67}\textcolor[HTML]{000000}{0.622} & \cellcolor[HTML]{15904C}\textcolor[HTML]{FFFFFF}{0.785} & \cellcolor[HTML]{FBA05B}\textcolor[HTML]{000000}{0.179} & \cellcolor[HTML]{91D068}\textcolor[HTML]{000000}{0.610} & \cellcolor[HTML]{5DB961}\textcolor[HTML]{000000}{0.685} & \cellcolor[HTML]{F67A49}\textcolor[HTML]{000000}{0.124} & \cellcolor[HTML]{9BD469}\textcolor[HTML]{000000}{0.595} & \cellcolor[HTML]{DE402E}\textcolor[HTML]{FFFFFF}{0.034} & \cellcolor[HTML]{FEE999}\textcolor[HTML]{000000}{0.320} \\
en-ro\_RO & \cellcolor[HTML]{7FC866}\textcolor[HTML]{000000}{0.634} & \cellcolor[HTML]{4BB05C}\textcolor[HTML]{000000}{0.707} & \cellcolor[HTML]{2AA054}\textcolor[HTML]{FFFFFF}{0.748} & \cellcolor[HTML]{128A49}\textcolor[HTML]{FFFFFF}{0.796} & \cellcolor[HTML]{249D53}\textcolor[HTML]{FFFFFF}{0.753} & \cellcolor[HTML]{89CC67}\textcolor[HTML]{000000}{0.619} & \cellcolor[HTML]{B7E075}\textcolor[HTML]{000000}{0.546} & \cellcolor[HTML]{78C565}\textcolor[HTML]{000000}{0.648} & \cellcolor[HTML]{AFDD70}\textcolor[HTML]{000000}{0.561} & \cellcolor[HTML]{1E9A51}\textcolor[HTML]{FFFFFF}{0.762} \\
en-ru\_RU & \cellcolor[HTML]{A5D86A}\textcolor[HTML]{000000}{0.580} & \cellcolor[HTML]{78C565}\textcolor[HTML]{000000}{0.647} & \cellcolor[HTML]{63BC62}\textcolor[HTML]{000000}{0.677} & \cellcolor[HTML]{36A657}\textcolor[HTML]{FFFFFF}{0.731} & \cellcolor[HTML]{4BB05C}\textcolor[HTML]{000000}{0.707} & \cellcolor[HTML]{BDE379}\textcolor[HTML]{000000}{0.534} & \cellcolor[HTML]{D1EC86}\textcolor[HTML]{000000}{0.499} & \cellcolor[HTML]{A7D96B}\textcolor[HTML]{000000}{0.575} & \cellcolor[HTML]{D1EC86}\textcolor[HTML]{000000}{0.500} & \cellcolor[HTML]{2DA155}\textcolor[HTML]{FFFFFF}{0.742} \\
en-sr\_Cyrl\_RS & \cellcolor[HTML]{51B35E}\textcolor[HTML]{000000}{0.699} & \cellcolor[HTML]{17934E}\textcolor[HTML]{FFFFFF}{0.775} & \cellcolor[HTML]{45AD5B}\textcolor[HTML]{000000}{0.714} & \cellcolor[HTML]{2DA155}\textcolor[HTML]{FFFFFF}{0.743} & \cellcolor[HTML]{33A456}\textcolor[HTML]{FFFFFF}{0.737} & \cellcolor[HTML]{A7D96B}\textcolor[HTML]{000000}{0.577} & \cellcolor[HTML]{A7D96B}\textcolor[HTML]{000000}{0.577} & \cellcolor[HTML]{73C264}\textcolor[HTML]{000000}{0.655} & \cellcolor[HTML]{6BBF64}\textcolor[HTML]{000000}{0.664} & \cellcolor[HTML]{54B45F}\textcolor[HTML]{000000}{0.696} \\
en-sr\_Latn\_RS & \cellcolor[HTML]{36A657}\textcolor[HTML]{FFFFFF}{0.731} & \cellcolor[HTML]{148E4B}\textcolor[HTML]{FFFFFF}{0.789} & \cellcolor[HTML]{3CA959}\textcolor[HTML]{FFFFFF}{0.724} & \cellcolor[HTML]{57B65F}\textcolor[HTML]{000000}{0.691} & \cellcolor[HTML]{128A49}\textcolor[HTML]{FFFFFF}{0.797} & \cellcolor[HTML]{66BD63}\textcolor[HTML]{000000}{0.672} & \cellcolor[HTML]{BFE47A}\textcolor[HTML]{000000}{0.532} & \cellcolor[HTML]{6EC064}\textcolor[HTML]{000000}{0.661} & \cellcolor[HTML]{ADDC6F}\textcolor[HTML]{000000}{0.564} & \cellcolor[HTML]{91D068}\textcolor[HTML]{000000}{0.610} \\
en-sv\_SE & \cellcolor[HTML]{6EC064}\textcolor[HTML]{000000}{0.662} & \cellcolor[HTML]{30A356}\textcolor[HTML]{FFFFFF}{0.738} & \cellcolor[HTML]{17934E}\textcolor[HTML]{FFFFFF}{0.777} & \cellcolor[HTML]{097940}\textcolor[HTML]{FFFFFF}{0.830} & \cellcolor[HTML]{16914D}\textcolor[HTML]{FFFFFF}{0.780} & \cellcolor[HTML]{7FC866}\textcolor[HTML]{000000}{0.634} & \cellcolor[HTML]{A9DA6C}\textcolor[HTML]{000000}{0.573} & \cellcolor[HTML]{4BB05C}\textcolor[HTML]{000000}{0.706} & \cellcolor[HTML]{7AC665}\textcolor[HTML]{000000}{0.641} & \cellcolor[HTML]{118848}\textcolor[HTML]{FFFFFF}{0.798} \\
en-th\_TH & \cellcolor[HTML]{0B7D42}\textcolor[HTML]{FFFFFF}{0.821} & \cellcolor[HTML]{05713C}\textcolor[HTML]{FFFFFF}{0.845} & \cellcolor[HTML]{07753E}\textcolor[HTML]{FFFFFF}{0.837} & \cellcolor[HTML]{69BE63}\textcolor[HTML]{000000}{0.667} & \cellcolor[HTML]{08773F}\textcolor[HTML]{FFFFFF}{0.831} & \cellcolor[HTML]{17934E}\textcolor[HTML]{FFFFFF}{0.775} & \cellcolor[HTML]{A2D76A}\textcolor[HTML]{000000}{0.585} & \cellcolor[HTML]{118848}\textcolor[HTML]{FFFFFF}{0.797} & \cellcolor[HTML]{7DC765}\textcolor[HTML]{000000}{0.639} & \cellcolor[HTML]{33A456}\textcolor[HTML]{FFFFFF}{0.735} \\
en-tr\_TR & \cellcolor[HTML]{4BB05C}\textcolor[HTML]{000000}{0.704} & \cellcolor[HTML]{219C52}\textcolor[HTML]{FFFFFF}{0.758} & \cellcolor[HTML]{45AD5B}\textcolor[HTML]{000000}{0.713} & \cellcolor[HTML]{75C465}\textcolor[HTML]{000000}{0.649} & \cellcolor[HTML]{15904C}\textcolor[HTML]{FFFFFF}{0.782} & \cellcolor[HTML]{89CC67}\textcolor[HTML]{000000}{0.619} & \cellcolor[HTML]{D1EC86}\textcolor[HTML]{000000}{0.498} & \cellcolor[HTML]{7AC665}\textcolor[HTML]{000000}{0.642} & \cellcolor[HTML]{C7E77F}\textcolor[HTML]{000000}{0.516} & \cellcolor[HTML]{30A356}\textcolor[HTML]{FFFFFF}{0.738} \\
en-uk\_UA & \cellcolor[HTML]{78C565}\textcolor[HTML]{000000}{0.646} & \cellcolor[HTML]{4BB05C}\textcolor[HTML]{000000}{0.704} & \cellcolor[HTML]{2AA054}\textcolor[HTML]{FFFFFF}{0.745} & \cellcolor[HTML]{1B9950}\textcolor[HTML]{FFFFFF}{0.763} & \cellcolor[HTML]{279F53}\textcolor[HTML]{FFFFFF}{0.752} & \cellcolor[HTML]{9BD469}\textcolor[HTML]{000000}{0.594} & \cellcolor[HTML]{B5DF74}\textcolor[HTML]{000000}{0.550} & \cellcolor[HTML]{7AC665}\textcolor[HTML]{000000}{0.643} & \cellcolor[HTML]{ABDB6D}\textcolor[HTML]{000000}{0.568} & \cellcolor[HTML]{18954F}\textcolor[HTML]{FFFFFF}{0.771} \\
en-vi\_VN & \cellcolor[HTML]{45AD5B}\textcolor[HTML]{000000}{0.714} & \cellcolor[HTML]{1E9A51}\textcolor[HTML]{FFFFFF}{0.762} & \cellcolor[HTML]{1E9A51}\textcolor[HTML]{FFFFFF}{0.762} & \cellcolor[HTML]{7DC765}\textcolor[HTML]{000000}{0.641} & \cellcolor[HTML]{097940}\textcolor[HTML]{FFFFFF}{0.827} & \cellcolor[HTML]{5DB961}\textcolor[HTML]{000000}{0.685} & \cellcolor[HTML]{CDEA83}\textcolor[HTML]{000000}{0.507} & \cellcolor[HTML]{51B35E}\textcolor[HTML]{000000}{0.698} & \cellcolor[HTML]{C5E67E}\textcolor[HTML]{000000}{0.522} & \cellcolor[HTML]{2DA155}\textcolor[HTML]{FFFFFF}{0.743} \\
en-zh\_CN & \cellcolor[HTML]{B1DE71}\textcolor[HTML]{000000}{0.557} & \cellcolor[HTML]{82C966}\textcolor[HTML]{000000}{0.633} & \cellcolor[HTML]{73C264}\textcolor[HTML]{000000}{0.653} & \cellcolor[HTML]{73C264}\textcolor[HTML]{000000}{0.653} & \cellcolor[HTML]{5AB760}\textcolor[HTML]{000000}{0.688} & \cellcolor[HTML]{A2D76A}\textcolor[HTML]{000000}{0.584} & \cellcolor[HTML]{C3E67D}\textcolor[HTML]{000000}{0.525} & \cellcolor[HTML]{A2D76A}\textcolor[HTML]{000000}{0.584} & \cellcolor[HTML]{C7E77F}\textcolor[HTML]{000000}{0.518} & \cellcolor[HTML]{2DA155}\textcolor[HTML]{FFFFFF}{0.744} \\
ja-zh\_CN & \cellcolor[HTML]{CBE982}\textcolor[HTML]{000000}{0.508} & \cellcolor[HTML]{B3DF72}\textcolor[HTML]{000000}{0.553} & \cellcolor[HTML]{70C164}\textcolor[HTML]{000000}{0.658} & \cellcolor[HTML]{33A456}\textcolor[HTML]{FFFFFF}{0.735} & \cellcolor[HTML]{16914D}\textcolor[HTML]{FFFFFF}{0.779} & \cellcolor[HTML]{7DC765}\textcolor[HTML]{000000}{0.639} & \cellcolor[HTML]{BFE47A}\textcolor[HTML]{000000}{0.532} & \cellcolor[HTML]{7DC765}\textcolor[HTML]{000000}{0.639} & \cellcolor[HTML]{B9E176}\textcolor[HTML]{000000}{0.545} & \cellcolor[HTML]{42AC5A}\textcolor[HTML]{000000}{0.718} \\
\bottomrule
\end{tabular}